# Stability analysis of tensegrity mechanism coupled with a bio-inspired piping inspection robot

S. Venkateswaran[a,*], and D. Chablat[b]

[a] *Léonard de Vinci Pôle Universitaire, Research Center, 92916 Paris La Défense, France* ;
[b] *Centre National de la Recherche Scientifique (CNRS), Laboratoire des Sciences du Numérique de Nantes (LS2N), UMR CNRS 6004, Nantes, 44321 France*

## 1. Introduction

Piping inspection robots play an essential role for industries as they can reduce human effort and pose a lesser risk to their lives. Generally, the locomotion techniques of these robots can be classified into mechanical and bioinspired (Kassim et al. 2006). By using slot-follower leg mechanisms, DC-motors, and control units, a rigid caterpillar type inspection robot was designed and developed at LS2N, France (Venkateswaran et al. 2019a). This rigid prototype helped in identifying the static forces required to accomplish good contact forces with the pipeline walls. In order to work inside curvatures, a tensegrity mechanism that uses three tension springs and a passive universal joint was introduced between each module of this robot (Venkateswaran et al. 2019b). The optimal parameters of the robot assembly were identified by considering a preload of the cables, which ensured the stability of the entire robot (Venkateswaran et al. 2021). However, under static conditions, there exist some forces on the robot, especially on the tensegrity mechanism when one end of the leg mechanism is clamped with the pipeline walls. These forces are dominant when the orientation of the pipeline is horizontal. The objective of this article is to understand the effect of the stiffness of the spring on the static stability of the tensegrity mechanism under the self-weight of the robot assembly.

## 2. Methods

Similar to the static force modelling approach followed by Venkateswaran et al., 2019a, the digital model of the robot is employed to analyse the tensegrity mechanism. The flexible robot in the horizontal configuration with one set of legs (left) fully clamped is shown in Figure 1. Under the condition proposed in Figure 1, the free end of the robot imposes forces and moments on the clamped end. As there are flexible tensegrity structures present between each module, the overall weight distribution

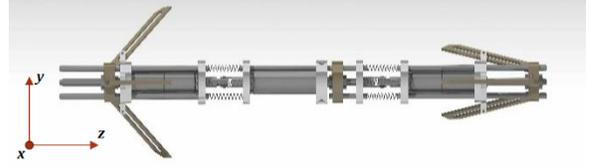

Figure 1 The piping inspection robot in the horizontal orientation

for the robot could be split as shown below in Figure 2.

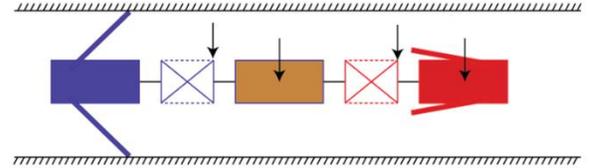

Figure 2 Distribution of masses on the robot under static condition

The tensegrity mechanism on the right end is subject to a higher amount of loading under the static condition as shown in Figure 2. The mechanism carries the weight of the entire robot, which is concentrated on the centre of gravity of the mechanism. While working inside a pipeline, the orientation of the robot about the z-axis is usually unknown. Based on earlier studies, the tensegrity mechanism undergoes the maximum deflection in the position when one of the springs is at a distance $r_f$ from the origin. This configuration will be taken for further analysis. The geometry of the mechanism in this orientation is shown below in Figure 3.

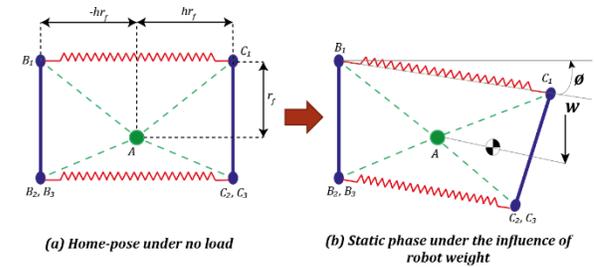

(a) Home-pose under no load  (b) Static phase under the influence of robot weight

Figure 3 The tensegrity mechanism in the home-pose and loading conditions

To recollect, the base coordinates of the mechanism are represented by $B_1$ to $B_3$. The moving platform coordinates are represented by $C_1$ to $C_3$. The springs are mounted at a distance $r_f$ from the origin of each platform. The origin of the mechanism and the universal joint is at $A$ and the tilt angles are given by $\alpha$ and $\beta$. When one set of leg mechanisms is clamped with the pipeline walls, the self-weight of the robot $w$ generates a deflection of the mechanism through an angle $\varnothing$. The self-weight is concentrated on the CG of the mechanism. The parameter $h$ determines the static stability of the mechanism under vertical configuration and it was found to be 0.6 (Venkateswaran et al. 2019b).

## 3. Results & Discussion

Under static conditions as demonstrated in Figure 2, the self-weight of the robot $w$ is estimated using the digital model of the robot in CATIA. The value was found to be around 4 N. As the mechanism is studied under a passive mode, the total potential energy of the system is contributed mainly by the springs ($U_s$) and the gravity ($U_g$). The total potential energy ($U_{tot}$) is given by:

$$U_{tot} = U_s + U_g \quad (1)$$

$$where\ U_s = \frac{1}{2} k \sum_{i=1}^{3}(l_i - l_o)^2$$

$$and\ U_g = w \sin \emptyset \ (r_f h + CG)$$

The parameter $l_o$ represents the zero free length of the spring and for ease of calculations, this is considered as 0mm. From the existing prototype (Venkateswaran et al. 2021), the value of $r_f$ was identified to be 11 mm. For the configuration represented in Figure 3b, the value of tilt angle α = 0 radians whereas β contributes to the tilt. At this configuration, the value of ø corresponds to the value of β. By solving Equation (1) using the design parameters specified, the total potential energy of the system is given by the equation:

$$U_{tot} = 35.2 \sin \beta + 312.8\ k - 50.82\ k \cos \beta \quad (2)$$

From Equation (2), it can be interpreted that the stiffness of the spring $k$ plays an essential role in the static stability of the mechanism. A value of 1 N/mm does not ensure a stable configuration. However, when the value reaches 20 N/mm, a stable potential energy configuration could be observed. A comparison of the potential energies for the two spring stiffness values is represented in Figure 4. From Figure 4b, it can also be concluded that the total potential energy of the system can never be a minimum at 0 radians. However, with the increase in spring stiffness, the deflection of the mechanism could be significantly reduced.

## 4. Conclusion

The analysis presented in this article helped in understanding the static stability of the tensegrity mechanism due to the self-weigh of the robot assembly. The effect of spring stiffness on the mechanism was also understood. This study also helps in modelling the stacked tensegrity structure that will be incorporated in the robot assembly and this will permit to have an elephant trunk design.

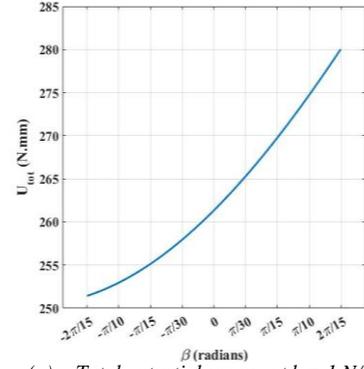

*(a)   Total potential energy at k = 1 N/mm*

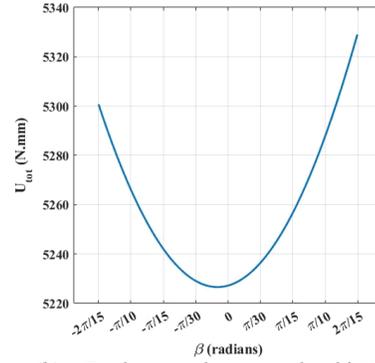

*(b)   Total potential energy at k = 20 N/mm*

Figure 4 Total potential energy of the tensegrity mechanism under varying stiffness of spring

*Corresponding author. Email:
swaminath.venkateswaran@devinci.fr